\documentclass{article}

\usepackage{graphicx}
\usepackage{float}          

\usepackage[preprint]{neurips_2025}

\usepackage[utf8]{inputenc} 
\usepackage[T1]{fontenc}    
\usepackage{hyperref}       
\usepackage{url}            
\usepackage{booktabs}       
\usepackage{amsfonts}       
\usepackage{nicefrac}       
\usepackage{microtype}      
\usepackage[dvipsnames, svgnames, x11names]{xcolor}

\definecolor{newgreen}{HTML}{11DDDD}
\definecolor{newpurple}{HTML}{B133E0}
\definecolor{newgrey}{HTML}{2A363B}
\definecolor{newolive}{HTML}{808000}
\definecolor{newcoral}{HTML}{FF7F50}

\usepackage[most]{tcolorbox}
\tcbset{
  red/.style={
    enhanced,
    colback=Salmon!20,
    colframe=Salmon!90!black,
    fonttitle=\small\bfseries,
    boxrule=0.8pt,
    rounded corners,
    left=0.15pt, right=0.15pt, top=0.05pt, bottom=0.05pt
  },
  green/.style={
    enhanced,
    colback=Emerald!10,
    colframe=newgreen!90!black,
    fonttitle=\small\bfseries,
    boxrule=0.8pt,
    rounded corners,
    left=0.15pt, right=0.15pt, top=0.05pt, bottom=0.05pt
  },
  purple/.style={
    enhanced,
    colback=MediumOrchid!10,
    colframe=newpurple,
    fonttitle=\small\bfseries,
    boxrule=0.8pt,
    rounded corners,
    left=0.15pt, right=0.15pt, top=0.05pt, bottom=0.05pt
  },
  blue/.style={
    enhanced,
    colback=SeaGreen!10!CornflowerBlue!10,
    colframe=RoyalPurple!55!Aquamarine!100!,
    fonttitle=\small\bfseries,
    boxrule=0.8pt,
    rounded corners,
    left=0.15pt, right=0.15pt, top=0.05pt, bottom=0.05pt
  },
  orange/.style={
   enhanced,
   colback=Orange!20,         
   colframe=Orange!90!black, 
   fonttitle=\small\bfseries,
   boxrule=0.8pt,
   rounded corners,
   left=0.15pt, right=0.15pt, top=0.05pt, bottom=0.05pt 
  },
  brown/.style={
   enhanced,
   colback=Brown!15,         
   colframe=Brown!80!black, 
   fonttitle=\small\bfseries,
   boxrule=0.8pt,
   rounded corners,
   left=0.15pt, right=0.15pt, top=0.05pt, bottom=0.05pt
  },
  cyan/.style={
   enhanced,
   colback=cyan!10,         
   colframe=cyan!90!black, 
   fonttitle=\small\bfseries,
   boxrule=0.8pt,
   rounded corners,
   left=0.15pt, right=0.15pt, top=0.05pt, bottom=0.05pt
  },
  grey/.style={
   enhanced,
   colback=newgrey!10,         
   colframe=newgrey, 
   fonttitle=\small\bfseries,
   boxrule=0.8pt,
   rounded corners,
   left=0.15pt, right=0.15pt, top=0.05pt, bottom=0.05pt
  },
  pink/.style={
   enhanced,
   colback=Pink!10,
   colframe=Pink!90!black, 
   fonttitle=\small\bfseries,
   boxrule=0.8pt,
   rounded corners,
   left=0.15pt, right=0.15pt, top=0.05pt, bottom=0.05pt
  },
  olive/.style={
   enhanced,
   colback=newolive!10,         
   colframe=newolive, 
   fonttitle=\small\bfseries,
   boxrule=0.8pt,
   rounded corners,
   left=0.15pt, right=0.15pt, top=0.05pt, bottom=0.05pt
  },
  coral/.style={
   enhanced,
   colback=newcoral!10,         
   colframe=newcoral, 
   fonttitle=\small\bfseries,
   boxrule=0.8pt,
   rounded corners,
   left=0.15pt, right=0.15pt, top=0.05pt, bottom=0.05pt
  },
}

\usepackage{listings}
\lstdefinestyle{python}{
    language = Python,
    backgroundcolor = \color{teal!10},  
    frame = shadowbox,                  
    numbers = left,                     
    morekeywords = {as}                 
}
\lstdefinestyle{pythonSimple}{
    language = Python,
    morekeywords = {as}                 
}
\lstdefinestyle{cpp}{
    language = C++,
    backgroundcolor = \color{teal!10},  
    frame = shadowbox,                  
    numbers = left                      
}
\lstdefinestyle{cppSimple}{
    language = C++
}
\lstdefinestyle{txt}{
    backgroundcolor = \color{teal!10},
    frame = shadowbox,
    basicstyle = \small\ttfamily
}
\lstdefinestyle{txtSimple}{
    basicstyle = \small\ttfamily
}

\usepackage{amsmath, color-edits, caption}
\addauthor[Yuhan]{yuhan}{blue}
\addauthor[Yimi]{yimi}{orange}

\title{Can LLMs Generate Reliable Test Case Generators? A Study on Competition-Level Programming Problems}

\author{%
  Yuhan Cao$^{1,2}$\footnotemark[1]\\
  \texttt{caoyh1@shanghaitech.edu.cn}
  \And
  Zian Chen$^{2}$\footnotemark[1]
  \And
  Kun Quan$^{3}$
  \And
  Ziliang Zhang$^{4}$
  \And
  Yu Wang$^{5}$
  \And
  Xiaoning Dong$^{6}$
  \And
  Yeqi Feng$^{1,7}$\footnotemark[2]
  \And
  Guanzhong He$^{6}$\footnotemark[2]
  \And
  Jingcheng Huang$^{8}$\footnotemark[2]
  \And
  Jianhao Li$^{2}$\footnotemark[2]
  \And
  Yixuan Tan$^{9}$\footnotemark[2]
  \And
  Jiafu Tang$^{8}$\footnotemark[2]
  \And
  Yilin Tang$^{1}$\footnotemark[2]
  \And
  Junlei Wu$^{2}$\footnotemark[2]
  \And
  Qianyu Xiao$^{10}$\footnotemark[2]
  \And
  Can Zheng$^{1}$\footnotemark[2]
  \And
  Shouchen Zhou$^{2}$\footnotemark[2]
  \And
  Yuxiang Zhu$^{2}$\footnotemark[2]
  \And
  Yiming Huang$^{11}$\\
  \And
  Tianxing He$^{6,1}$\footnotemark[3]\\
  \texttt{hetianxing@mail.tsinghua.edu.cn}\\
  $^1$ Shanghai Qi Zhi Institute
  \quad
  $^2$ ShanghaiTech University
  \quad
  $^3$ Wuhan University\\
  $^4$ Fuzhou University
  \quad
  $^5$ Institute of Information Engineering, Chinese Academy of Sciences\\
  $^6$ Tsinghua University
  \quad
  $^7$ Huazhong University of Science and Technology\\
  $^8$ Nanjing University
  \quad
  $^{9}$ Beijing University of Posts and Telecommunications\\
  $^{10}$ Peking University
  \quad
  $^{11}$ Independent Researcher\\
}

\hypersetup{hidelinks}

\begin{document}

\renewcommand{\thefootnote}{\fnsymbol{footnote}}
\footnotetext[1]{Co-first authors.}
\footnotetext[2]{Equal contribution. Authors are listed alphabetically.}
\footnotetext[3]{Corresponding author.}

\maketitle

\renewcommand{\thefootnote}{\arabic{footnote}}

\begin{abstract}
  Large Language Models (LLMs) have demonstrated remarkable capabilities in code generation, capable of tackling complex tasks during inference. However, the extent to which LLMs can be utilized for code checking or debugging through test case generation remains largely unexplored. We investigate this problem from the perspective of competition-level programming (CP) programs and propose \textbf{TCGBench}, a \textit{Benchmark for (LLM generation of) Test Case Generators}. This benchmark comprises two tasks, aimed at studying the capabilities of LLMs in (1) generating valid test case generators for a given CP problem, and further (2) generating targeted test case generators that expose bugs in human-written code. Experimental results indicate that while state-of-the-art LLMs can generate valid test case generators in most cases, most LLMs struggle to generate targeted test cases that reveal flaws in human code effectively. Especially, even advanced reasoning models (e.g., o3-mini) fall significantly short of human performance in the task of generating targeted generators. Furthermore, we construct a high-quality, manually curated dataset of instructions for generating targeted generators. Analysis demonstrates that the performance of LLMs can be enhanced with the aid of this dataset, by both prompting and fine-tuning. Code is available at \url{https://github.com/TCGBench/TCGBench}.
\end{abstract}













\section{Introduction}

Code generation is one of the most important tasks for large language models (LLMs). With the advancement of model performance, LLMs have demonstrated remarkable capabilities across various code-related tasks \citep{doi:10.1126/science.abq1158,jiang2024surveylargelanguagemodels}. The common method for evaluating code generation tasks involves testing with test cases, primarily because code correctness is typically determined through execution. Consequently, the ability to find and correct code errors is an important skill for human coders. These errors are not only syntax errors but also need reasoning based on code semantics or specific details. Therefore, verifying code correctness by generating test cases holds significant importance.

This leads us to consider: do models possess the capability to verify code correctness by generating test cases? We approach this question from the perspective of competition-level programming (CP) problems. These problems are effective for testing a model's reasoning ability in relatively short contexts and have been widely used since they are crafted by experts for wide-ranging competitions and education. In parallel with the evolution of reasoning models, a growing body of work has begun to focus on evaluating language model capabilities through their use \citep {jain2025livecodebench,deepseekai2025deepseekv3technicalreport,qwen2025qwen25technicalreport}. We observe that within these evaluation efforts \citep{jain2025livecodebench,kimiteam2025kimik15scalingreinforcement}, researchers utilize LLMs to automatically generate test case generators. However, the process for generating generators is either overly simplistic or lacks a thorough analysis regarding the reliability of the data generated by the models.

On the other hand, while advanced reasoning models have demonstrated significant capabilities on CP problems, some work \citep{huang2024competitionlevelproblemseffectivellm} suggests that models may exhibit a tendency towards memorizing solutions rather than performing true reasoning. The extent to which models genuinely comprehend the semantics of code also represents a question warranting attention. Furthermore, the precise scope of language models' generation capabilities in this domain remains under-researched. 

To systematically investigate this problem, we propose a novel benchmark, TCGBench, a \textit{Benchmark for (LLM generation of) Test Case Generators}. The contributions of our work are as follows:

\begin{itemize}
    \item We collect 208 problems from two representative CP problem data sources, along with over a thousand corresponding standard solver programs and erroneous programs. Utilizing this data, we propose two evaluation tasks designed to evaluate the capability of models in generating test case generators, specifically evaluating their ability to produce valid generators and targeted generators.
    \item We create a carefully curated, high-quality dataset of code analysis for writing targeted generators. Our analysis indicates that LLMs can achieve improved performance by using this dataset for prompting and fine-tuning.
\end{itemize}

\begin{figure}[ht]
  \centering
  \begin{tcolorbox}[blue,title=Problem Statement]
    Implement an Aho-Corasick (AC) automaton. Given a text string $S$ and $n$ pattern strings $T_{1 \sim n}$. For each pattern string $T_i$, determine how many times it appears in $S$.

    For $100\%$ of the data, $1 \le n \le 2 \times {10}^5$, the total length of $T_{1 \sim n}$ does not exceed $2 \times {10}^5$, and the length of $S$ does not exceed $2 \times {10}^6$.
  \end{tcolorbox}
  \begin{tcolorbox}[cyan, title=Valid Test Case Generator]
    \begin{lstlisting}[style=pythonSimple]
while len(patterns) < n and current_length < total_length_T:
    pattern_length = random.randint(1, min(20, total_length_T - current_length))
    pattern = generate_randomstring(pattern_length)
    patterns.append(pattern)
    current_length += pattern_length
S = generate_string(S_length)
inputs = [str(n)] + patterns + [S]
    \end{lstlisting}
  \end{tcolorbox}
  \begin{tcolorbox}[purple, title=Erroneous Solver Program]
    \begin{lstlisting}[style=cppSimple]
struct AhoCorasick
{
    std::vector<Node> t; // An AC automaton has several nodes.
    AhoCorasick(){
        init();
    }
    void init() {
        t.assign(2, Node()); 
        t[0].next.fill(1); // next: the links in the AC automaton
        t[0].len = -1;// len: the length of the string 
        // that maps with the path from the root to the current node.
    }// ...
    \end{lstlisting}
  \end{tcolorbox}


  \begin{tcolorbox}[orange,title=Target Instruction]
The core issue in the implementation stems from incorrectly initializing the root node's transition table to point to node 1, rather than 0. ...

To generate a test case that makes the code fail, we need to create input where:

1. Use pattern strings where the short patterns are the suffix or prefix of the longer ones. (e.g. \texttt{a},\texttt{ab},\texttt{cab},\texttt{abc})

2. The text string contains repeated overlaps of these patterns.
  \end{tcolorbox}
  \begin{tcolorbox}[green,title=Successful Targeted Generator]
    \begin{lstlisting}[style=pythonSimple]
for _ in range(n):
    length = random.randint(1, 10)  # Pattern length ranging from 1 to 10
    pattern = ''.join(random.choices(string.ascii_lowercase, k=length))
    patterns.append(pattern)
S_length = 2000000  # Close to the limit
S = ''.join(random.choices(patterns, k=S_length // 10))
inputs = []
inputs.append(str(n))
inputs.extend(patterns)
inputs.append(S)
    \end{lstlisting}
  \end{tcolorbox}
  \caption{An example of our benchmark. It consists of two tasks: generation of valid test case generators and generation of targeted test case generators. To enhance clarity, the example has been presented in a simplified format. For the full example, please refer to Appendix~\ref{sec:app_examples}.}
  \label{fig:example}
\end{figure}

\section{Benchmark Building}

\vspace{-1.5mm}


\subsection{Problem Data Source}

We use competition-level programming (CP) problems as our data source. These problems are effective for testing a model's reasoning ability in relatively short contexts and have been widely used since these problems are crafted by experts for wide-ranging competitions and education, and the correctness of code generated by LLMs is easy to verify.

We select two representative problem sets, utilized to train human competitive programmers, as our data source. Our data comes from two representative problem sets used to train programmers for competitions.



\paragraph{The NOIP Problem Set.} The NOIP Problem Set includes problems from the National Olympiad in Informatics in Provinces (NOIP), the contest of China’s National Olympiad in Informatics for adolescents. The term `province' in the name of this problem set indicates that the associated competition is conducted simultaneously nationwide, organized on a provincial basis, and utilizes identical test problems across the country. The problems in this competition are often inspired by real-life scenarios or well-defined mathematical tasks, with relatively lower difficulty in comprehension. We select the NOIP problems from 1998 to 2022 to serve as our initial problem set.


\paragraph{The Canonical Problem Set.} The Canonical Problem Set includes various algorithms and data structures implementation tasks. The tasks presented by these problems often involve the canonical implementation of target algorithms or data structures. They feature concise and clear problem statements, typically serving as the first exercise for human coders learning a particular algorithm or data structure. Following the common practice within the competitive programming community, we select all canonical problems on a third-party online judge (OJ) platform, Luogu\footnote{https://www.luogu.com.cn/}, to serve as our initial problem set.

All selected problems are publicly licensed to ensure compliance with research ethics. 

Since the focus of this work is test case generation, we want to first ensure that the LLMs can properly understand the problems. From the complete pool of NOIP and canonical problems, we first test with OpenAI o1-mini \citep{o1-mini}, only keeping the problems that o1-mini could fully pass at 1 attempt on Luogu. We then manually select those with clearly defined problem descriptions, data constraints, and a sufficient number of human-written codes. As a result, our final dataset consists of 129 NOIP problems and 79 canonical problems.





\subsection{Collection of Human-Written Code}

To build our benchmark, we require two types of code for evaluation. 

First, we need the correct reference solver programs. These programs are essential for validating the correctness of a test case generator, because fully determining a generator's validity through human expertise is difficult and time-consuming. Instead, we can utilize correct programs to perform a more effective verification by comparing the consistency of their outputs for valid inputs. We collect programs that successfully pass all test cases on Luogu for all problems in our problem set, which we refer to as \textbf{standard solver programs}. We manually filter them, retaining only C++ code for subsequent evaluation and ensuring that each problem has more than 5 standard solver programs. 

Furthermore, we also need to evaluate the ability of LLMs to understand and reason about errors inherent in code and generate a targeted test case generator. For this purpose, we collect code submissions that fail to pass all test cases on the OJ, which we refer to as \textbf{erroneous programs}. We further categorize and analyze these `errors' in more detail: In terms of evaluation results on the test case from the OJ platform, the results for these programs may correspond to one or more categories listed in Table~\ref{tab:oj_results}. To facilitate evaluation and ensure the code is related to the problem (otherwise the language model may confused on finding errors), we choose 5 erroneous programs that only pass the online judge test cases at a ratio in the range [0.6,1), and their errors consist solely of one or more from $\{\texttt{WA}, \texttt{TLE}, \texttt{RE}\}$.




\begin{table}[h]
    \centering
    \caption{Possible judge results.}
    \label{tab:oj_results}
    \begin{tabular}[h]{c|l}
        \toprule
        \multicolumn{2}{l}{\textbf{Judge Results}} \\
        \midrule
         Result & Description \\
        \midrule
        \texttt{AC} & Accepted, the solver program is correct. \\
        \texttt{WA} & Wrong Answer, the output of the solver program differs from the expected output. \\
        \texttt{TLE} & Time Limit Exceeded, the solver program took more time than allowed. \\
        \texttt{RE} & Runtime Error, the solver program encountered an error during execution. \\
        \texttt{MLE} & Memory Limit Exceeded, the solver program used more memory than allowed. \\
        \texttt{CE} & Compilation Error, the solver program failed to compile. \\
        \bottomrule
    \end{tabular}
\end{table}


\subsection{Target Instruction Dataset}

Finally, to systematically evaluate the capability of language models to generate targeted test case generators given erroneous programs and their errors (a detailed definition will be provided subsequently), we curate a subset of 66 erroneous programs from the dataset corresponding to the NOIP problem set. For each program in this subset, we meticulously annotate its error analysis, instructions for generating a targeted generator, and an example targeted generator. This dataset can be applied for model fine-tuning, in an effort to enhance the model's ability to generate targeted generators. Detailed examples are provided in Appendix~\ref{sec:app_ti_dataset}.

\section{Evaluating LLM's Ability in Generating Test Case Generators}

\vspace{-1.5mm}

\subsection{Terminologies and Notations}

In this section, we introduce the terminologies used in the construction of our benchmark.


In our benchmark, we denote a programming problem as a pair $(\pi, C)$, where $\pi$ is the problem statement, and $C$ represents a test case set (i.e., test suite in software engineering). The problem statement $\pi$ specifies the test case constraints for all variables within the test case input. 
Each test case $c_i=(\tau_i,\phi_i)$ contains a valid input $\tau_i$ as well as an expected output $\phi_i$. 

In the context of real-world algorithmic competitions, there might be multiple valid correct outputs for an input; however, for simplicity in evaluation, we will only consider problems where there is only one correct answer for any given test case input. Typically, the problem statement $\pi$ in CP problems specifies the variable constraints for all variables within the test case input\footnote{For example, a problem statement might include the following description: ``For all test cases, $1\le m\le 10^5$.'' It provides a variable constraint of $m$ in natural language.}.

A test case input generator $\Gamma$ is a program that generates random or deterministic test case inputs. In our benchmark, all generators are implemented in Python.


A solver program $\Sigma$ is a program that attempts to solve a given problem. It can be executed with a test case input and return an output. An automatic judge $J: (\tau,\phi,\Sigma)\to\omega$ can run a solver program, and return a judge result $\omega\in\{\texttt{AC},\texttt{WA}, \texttt{TLE}, \texttt{RE}\}$ concerning the test case $\tau$. Building upon this, we define a solver program as a \textbf{standard solver program} $\Sigma_{\text{std}}$ if it executes successfully and produces the expected output within a reasonable time for all test case inputs that satisfy the constraints. Correspondingly, a solver program is defined as a \textbf{erroneous solver program} $\Sigma_{\text{err}}$ if there exists at least one valid test case for which the program either fails to executes successfully or does not produce the expected output. This definition can be formalized as follows:

\begin{align}
    \forall (\tau,\phi)\in\mathit{T},\, J(\tau,\phi,\Sigma_{\text{std}})&=\texttt{AC},\\
\exists (\tau_e,\phi_e)\in\mathit{T},\, J(\tau_e,\phi_e,\Sigma_{\text{err}})&\in\{\texttt{WA},\texttt{RE},\texttt{TLE}\}.
\end{align}

\subsection{Task: The Generation of Valid Test Case Generators}

Now we formalize our main task: LLM generation of a test case generator. 


In this work, the prompt received by the LLM consists of the problem statement $\pi$ and the task instructions $\Theta$. The expected response is a test case generator $\Gamma$.

In this task, given a prompt, we have the language model generate $n$ samples of generators. For the generator $\Gamma_i$, we run it $m$ times to get $m$ test cases, and form these $m$ test cases as a new test case input set $T^{\text{in}}_i$. We evaluate the test case input set in two different ways and get a metric $r$.



Our first goal is to evaluate the model's ability to generate valid generators given a CP problem. In this task, the prompt to LLMs includes the problem statement and an instruction prompt. The LLM is instructed to generate a valid generator adhering to a specified format, referencing the problem statement and variable constraints. It also includes a Chain-of-Thought \citep{wei2022chain} prompt and a one-shot demonstration of a $(\pi, \Gamma)$ pair. In addition, we also manually fix the values of certain variables based on the problem description. They are mainly in the statements expressing the data ranges of quantities in the problem. We choose the largest possible value as the constraint. For example, a statement `Alice has $n\le10^5$ apples' indicates we have a constraint that $n=10^5$. 

The LLMs are required to set these variables to our specified values in their generators. The main purpose of this requirement is to prevent the LLM from ``cheating'' by generating generators with a tiny test case range, which could make the task trivial. We perform independent sampling from LLMs $n=10$ times, uniformly setting the temperature to 1 to get 10 generator samples. 

For each generator, we evaluate its validity by the standard solver programs. Specifically, we run the generator $m=10$ times to generate $m$ test cases. These test cases are then provided to at least 5 standard solver programs associated with the problem. For a given test case $\tau$, it is valid if all solver programs produce the same output and none of them crash or exceed the time limit of 10 wall-clock seconds. 

We define $r^{\pi}_i$ as an indicator variable, namely if generator $\Gamma_i$ is valid for problem $\pi$, $r_i=1$, otherwise $r_i=0$. The final metric $r^{\pi}$ is denoted as valid@k, which is formalized analogous to the popular pass@k metric, i.e.,

\begin{align}
r^{\pi}&=1-\frac{\binom{n-c}{k}}{\binom nk},
\end{align}

where $c=\sum_{i=1}^n[r_i=1]$.

\subsection{Evaluating Generation of Targeted Generators}
\label{subsec:task_target}

Our second goal is to evaluate the model's ability to generate targeted generators for specified erroneous solver programs. In this task, the instruction prompt is followed by an erroneous solver program $\Sigma_{\text{err}}$. The LLM is instructed to generate a targeted generator, which can generate a test case to expose flaws of that program, leading it to produce a wrong output, crash, or exceed the time limit. As an intermediate step, we also instruct the language model to explicitly output a reasoning process that covers code issues and instructions for writing the corresponding targeted generator. This serves as a form of Chain-of-Thought prompting. We also give a one-shot demonstration of a $(\pi, \Sigma_{\text{err}}, \Gamma)$ triplet.

Consistent with the first task, we perform independent sampling from LLMs $n=10$ times, uniformly setting the temperature to 1 for each sample. For each generator $\Gamma_i$, our requirements in this task are stricter. Specifically, for each $(\pi,\Sigma_{\text{err}})$ pair, the generator generates a single test case $\tau$ ($m=1$). This test case is then submitted both to a selected standard solver program ($\Sigma_{\text{std}}$) and the erroneous program ($\Sigma_{\text{err}}$). A targeted generator $\Gamma_i$ is considered successful if $\Sigma_{\text{std}}$ executes successfully, and the output of $\Sigma_{\text{err}}$ differs from that of $\Sigma_{\text{std}}$, or $\Sigma_{\text{err}}$ crashes or exceeds the time limit.

Similar to the first task, we define $r_i$ as an indicator variable, and the final metric $r^{(\pi,\Sigma_{\text{err}})}$ is denoted as success@k, formulated as: 

\begin{align}
r^{(\pi,\Sigma_{\text{err}})}&=1-\frac{\binom{n-c}{k}}{\binom nk},
\end{align}

where $c=\sum_{i=1}^n[r_i=1]$. In our benchmark, for each problem, we choose 5 corresponding erroneous solver programs.



\subsection{Human Performance as a Reference}

To better evaluate the performance of the language model on our task, we also collect data on human performance, serving as a reference.

6 experienced volunteers (all of them are undergraduate students with competitive programming experience of at least 3 years) are recruited to complete the two tasks stated above. For the first task, human experts are each assigned a total of 25-40 problems, with the requirement to produce a valid generator for each within a 15-minute time limit. For the second task, experts are each assigned a total of 50-80 erroneous solver programs, derived from the same 25-40 problems (with at least two erroneous programs per problem). For both tasks, failure is recorded if a generator is not successfully produced within this time constraint.

\section{Experimental Results}
\label{sec:results}

\vspace{-1.5mm}

\subsection{Valid Test Case Generator Generation}

\begin{figure}
    \centering
    \includegraphics[width=1\linewidth]{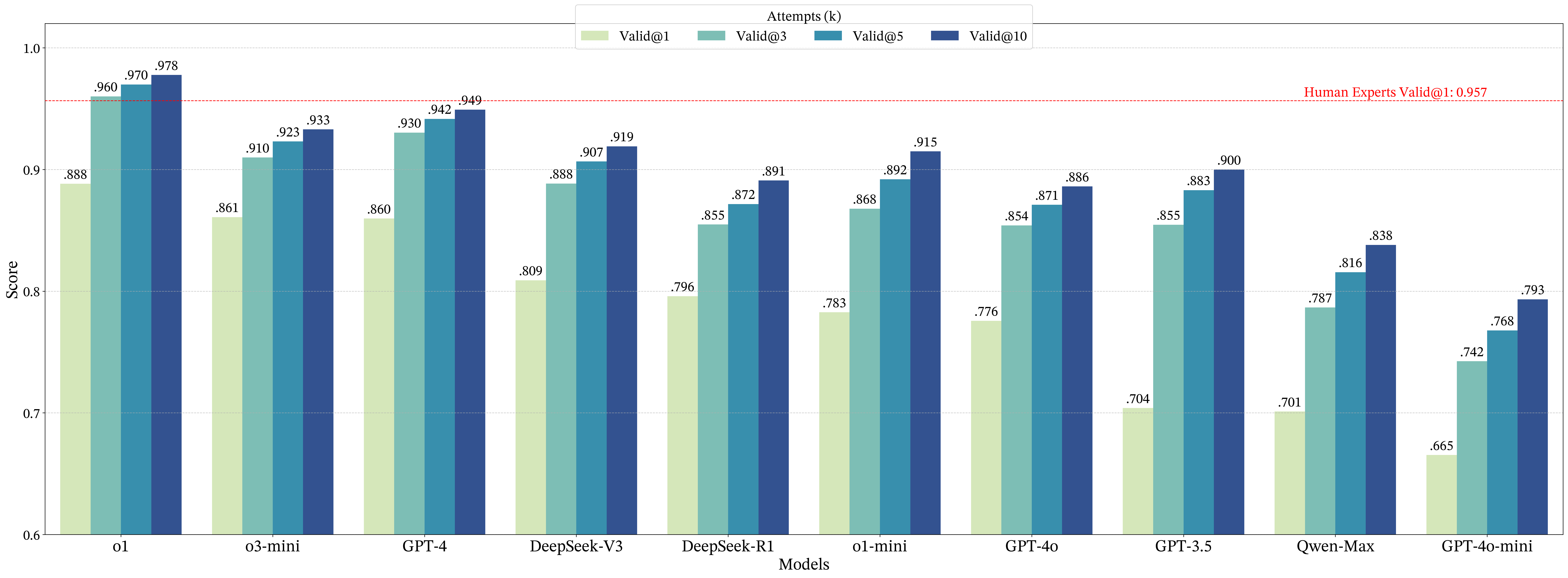}
    \caption{Valid@k results. When $k\ge5$, o1 demonstrated capabilities exceed human level. However, at $k=1$, the capabilities of most models still fall short of human level.}
    \label{fig:basic}
\end{figure}
\begin{figure}
    \centering
    \includegraphics[width=1\linewidth]{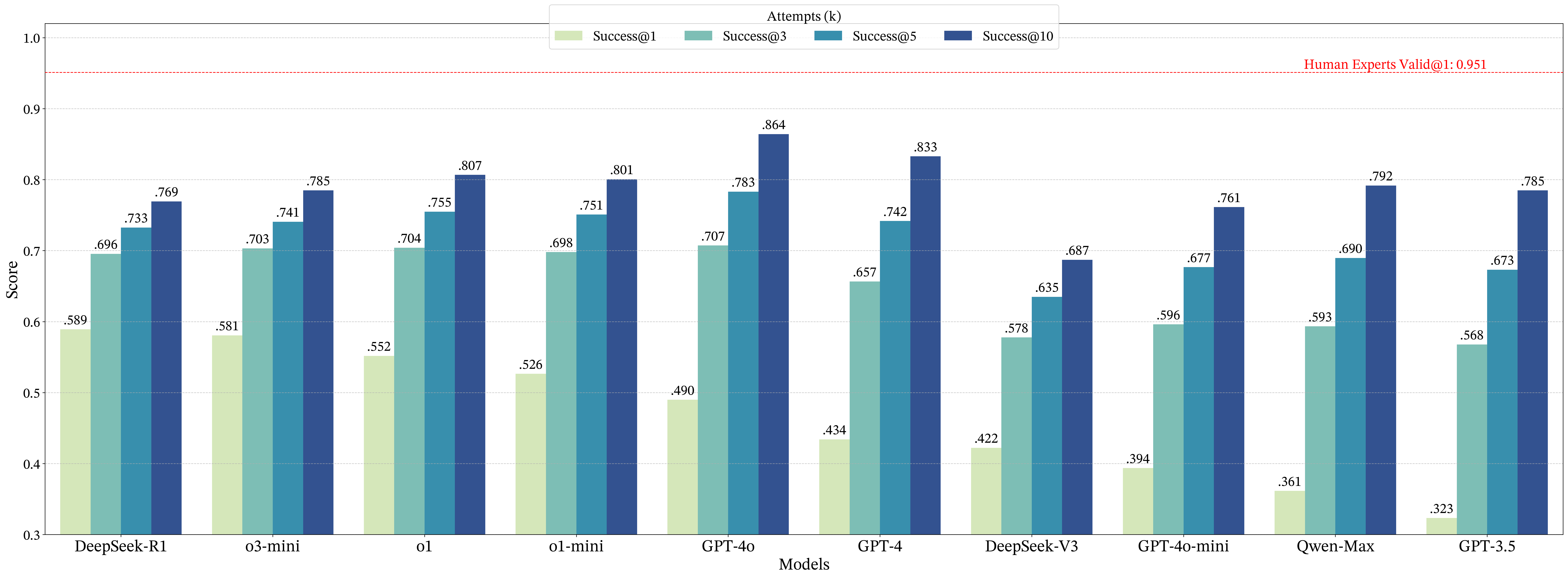}
    \caption{Success@k results. Most models have a poor performance at success@1. However, when the number of samples increases, success@k shows a significant improvement. This suggests that a decent level of ability to generate targeted generators, but requires multiple tries.}
    \label{fig:hack}
\end{figure}

In this section, we investigate our first research question regarding the model's ability to generate valid test case generators. We perform experiments on multiple state-of-the-art language models, including o3-mini \citep{o3-mini}, o1 \citep{openai2024openaio1card},  GPT-4 \citep{openai2024gpt4technicalreport}, GPT-4o series \citep{openai2024gpt4ocard,gpt-4o-mini}, Deepseek V3 \citep{deepseekai2025deepseekv3technicalreport}, Qwen-Max \citep{qwen2025qwen25technicalreport}. We find that, for this task, all models' performances on the two datasets are similar. Therefore, we present the results of the combined dataset, depicted in Figure~\ref{fig:basic}. The full results are detailed in Appendix~\ref{sec:app_tabular_results}.

As illustrated, the majority of models exhibit relatively strong performance on this task, especially with a greater number of attempts. Advanced reasoning models, such as o1, perform better. However, it is important to note that if the model fails to achieve a high valid@1 in this task (valid@1 of most models are above 0.7), this has certain implications for real-world scenarios, as in practical competitions, one invalid test case could potentially require significant costs for re-judging.



\subsection{Targeted Test Case Generator Generation}

Next, we present the results of our study on the targeted test case generator task. For ease of presentation, we similarly show the results from the combined dataset, as illustrated in Figure~\ref{fig:hack}. The full results are also available in Appendix~\ref{sec:app_tabular_results}.

We find that most models exhibit poor performance on the success@1 metric (success@1 of most models is below 0.6). Even state-of-the-art models show a significant capability gap compared to human experts. This disparity indicates that the models' ability to understand program errors is still quite limited.

Furthermore, it is worth noting that, in our dataset, many errors in the erroneous programs are relatively straightforward but involve specific details of the problem statement. Human experts can solve these types of issues with ease, yet LLM fails to produce the correct targeted generator. We provide an example illustrating this situation in Figure~\ref{fig:example}, and more examples in Appendix~\ref{sec:app_examples}.

From this evaluation result, we believe that addressing how to enable models to understand code-level details (as opposed to summarizing functionality, at which models are good) and identify the corresponding errors is a problem worthy of significant attention and future research.

\subsection{Curated Instructions Help LLMs Generate Targeted Generators Better}

\begin{figure}
    \centering
    \includegraphics[width=1\linewidth]{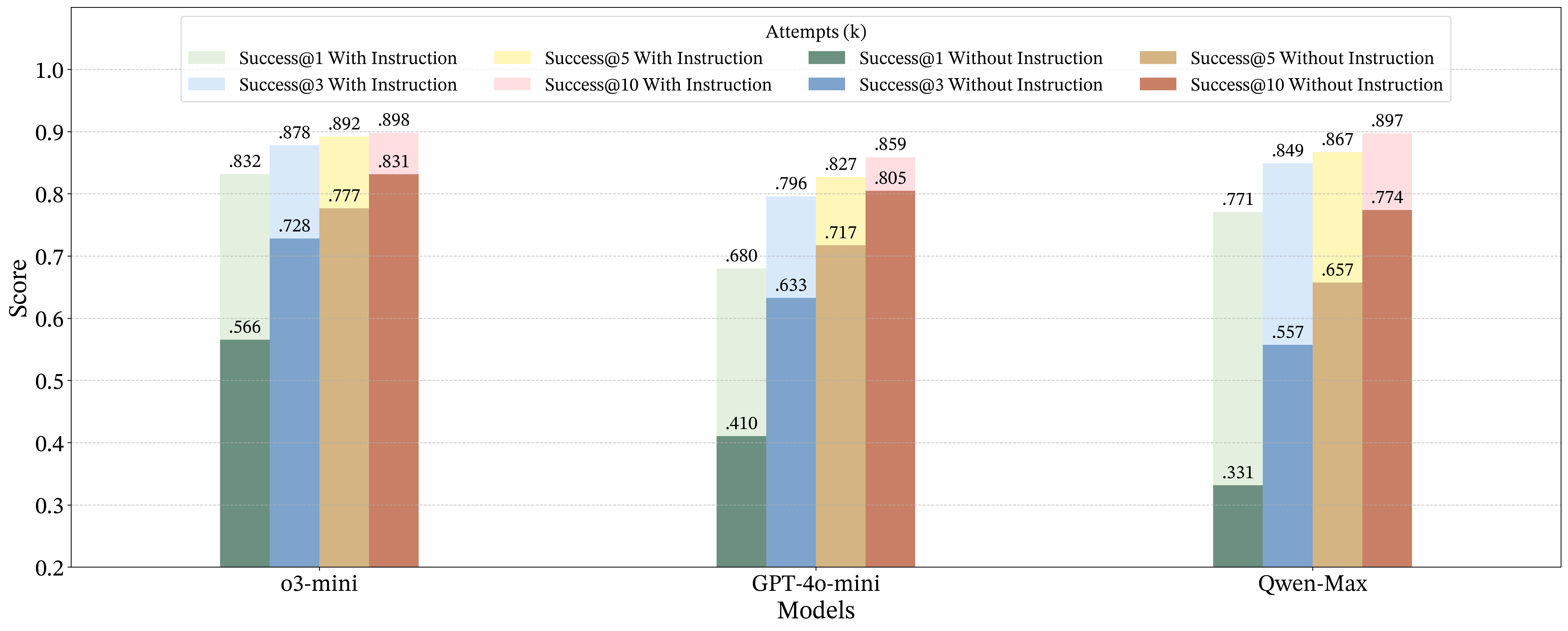}
    \caption{Success@k in Targeted Instruction Dataset. With the help of the target instructions, all 3 models show a significant improvement in success@1.}
    \label{fig:target}
\end{figure}

Finally, we analyze the impact of targeted instructions on model performance. First, we examine the application of our targeted instruction dataset as contextual information for advanced closed-source models. For each pair of problem and erroneous program(s) $(\pi, \Sigma_{\text{err}})$, we incorporate the target instruction (refer to Figure~\ref{fig:example}) associated with $\Sigma_{\text{err}}$ into the prompt, directing the language model to generate targeted generators based on the provided instruction. We select three representative models to illustrate our experimental findings, which are presented in Figure~\ref{fig:target}. Comprehensive experimental results are detailed in Appendix~\ref{sec:app_tabular_results}.

We observe that the presence of instructions leads to a significant improvement in the model's success@1 score. This suggests that our target instructions are highly beneficial in assisting the model's comprehension of the specific issues within erroneous programs. However, it is observed that the model could not generate targeted generators for all erroneous solver programs even after 10 sampling attempts, indicating that its performance on this task is constrained by its inherent capabilities.




On the other hand, we perform supervised fine-tuning (SFT) of the 14B variant of Qwen2.5 instruct model on the target instruction dataset using Low-Rank Adaptation (LoRA) \citep{hu2021loralowrankadaptationlarge}. The model was optimized using paged AdamW 8bit \citep{loshchilov2019decoupledweightdecayregularization} with a learning rate of $2\times10^{-4}$, weight decay of 0.01, and linear learning rate scheduling, trained for 100 epochs. All training samples were formatted with the Alpaca-style \citep{alpaca} instruction templates. We do experiments on the Canonical Problem Set, which are presented in Figure~\ref{fig:finetune}. Through a case study (detailed in Appendix~\ref{sec:app_examples}), we reveal that the base model exhibits severe hallucination. However, after fine-tuning, the model's output was effectively constrained to the correct format. We also find that the finetuned model tends to over-identify common issues in the given erroneous code snippets. This manifests in two key patterns:


\paragraph{Over-sensitivity to common mistakes.} The finetuned model frequently flags integer overflow or array out-of-bounds issues in the code snippets, even when safeguards exist. 

\paragraph{Misestimates time complexity.}
The finetuned model sometimes misestimates algorithmic time complexity, regardless of the given data range.

This might imply that the model can learn knowledge from the dataset, but due to its own capability, the improvement may be limited.

\begin{figure}
    \centering
    \includegraphics[width=1\linewidth]{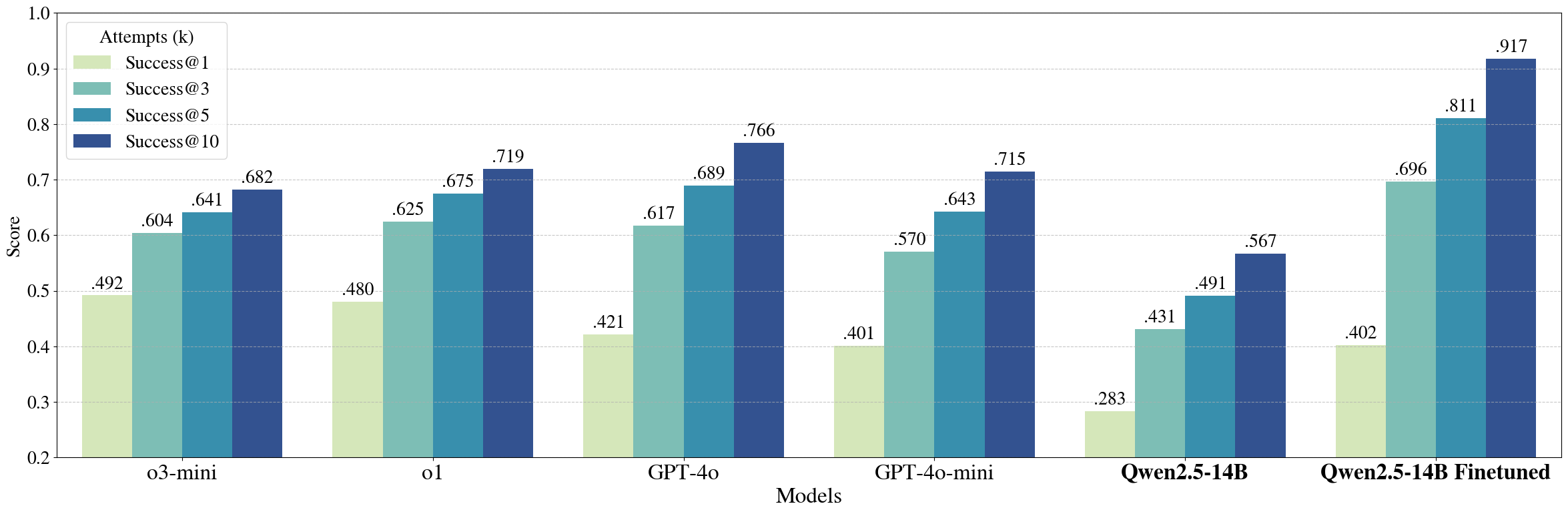}
    \caption{Success@k in the Canonical Problem Dataset. The last two groups show the results before and after fine-tuning Qwen2.5-14B. It can be observed that the fine-tuned Qwen2.5-14B achieves capabilities close to GPT-4o-mini.}
    \label{fig:finetune}
\end{figure}

\section{Related Work}

\vspace{-1.5mm}

\subsection{LLM for Code Generation}
Large Language Models (LLMs) have demonstrated remarkable capabilities in code generation. The success of prompt engineering techniques such as Chain-of-Thought \citep{wei2022chain} has been verified in numerous fields, including code generation. Structured Chain-of-Thought \citep{Li2023StructuredCP} explicitly incorporates programming structures (sequence, branch, and loop) into the reasoning process, guiding LLMs to think systematically about code design before implementation. ClarifyGPT \citep{Mu2024ClarifyGPTAF} introduces an interactive clarification mechanism that detects ambiguous requirements through code consistency checks, generates targeted clarifying questions, and refines requirements before final code generation.


Self-Debugging teaches LLMs to perform rubber duck debugging without external feedback, where models identify mistakes by explaining generated code line-by-line or tracing execution. LDB \citep{Zhong2024LDBAL} segments programs into basic blocks and tracks intermediate variable values during runtime execution, enabling LLMs to verify correctness against task descriptions block by block. SelfEvolve \citep{Jiang2023SelfEvolveAC} implements a two-step pipeline where LLMs extract knowledge from input prompts to generate intermediate code, then act as expert programmers to debug based on interpreter feedback without requiring special test cases. Reflexion \citep{Shinn2023ReflexionLA} learns through verbal feedback and self-reflection, with an agent explicitly critiquing each response based on external data, generating citations, and enumerating missing aspects of the response.

\subsection{LLM Capabilities on Competition-Level Programming Problems}


Beyond standard software development, competitive programming represents another significant domain for code generation tasks, yielding a vast and diverse collection of code snippets and programs. Competition-level problems are characterized by their high reasoning complexity and the demand for precise algorithmic implementation, thereby constituting a highly integrative reasoning challenge.

AlphaCode~\citep{li2022competition} was the first to introduce a pipeline for generating code in the context of competitive programming. Although Large Language Models (LLMs) have demonstrated proficiency in generating code for general tasks, the challenge becomes more significant when applied to this domain. The ability of LLMs to generate human-level solutions in such contexts has sparked discussions about their potential as general problem solvers. However, findings from the USACO benchmark~\citep{shi2024can} highlight the substantial challenges that LLMs currently face. This benchmark, featuring 307 problems from the USA Computing Olympiad that require complex algorithmic reasoning and puzzle-solving skills, reveals that GPT-4~\citep{openai2024gpt4technicalreport} achieves only 8.7\% pass@1 accuracy with zero-shot prompting. Even when augmented with methods like self-reflection and episodic retrieval, the model's accuracy improves to only 20.2\%, indicating persistent difficulty in comprehensively solving the benchmark. 

Nevertheless, with the advancement of large reasoning models, their performance on competition-level programming problems has been continuously improving. For instance, \citet{openai2025competitiveprogramminglargereasoning} systematically investigated the capabilities of these models in solving highly challenging competition-level programming problems. Concurrently, \citet{quan2025codeelobenchmarkingcompetitionlevelcode} has benchmarked the performance of various language models within a general, human-comparable setting.

\section{Conclusion}



This work underscores the critical importance of equipping large language models with the capability to verify code correctness, particularly through the generation of reliable test cases. While LLMs have shown impressive strides in code generation, the equally vital task of verifying their output and that of human-written code remains a significant challenge. Our investigation, grounded in the context of competitive programming, revealed limitations in existing evaluation methodologies concerning the systematic assessment and analysis of LLM-generated test case generators. Furthermore, open questions persist regarding the extent of LLMs' true reasoning abilities versus mere memorization and their genuine comprehension of code semantics.

To address these gaps and provide a rigorous framework for evaluation, we introduce TCGBench, a novel benchmark specifically designed for assessing the capability of LLMs to generate test case generators. Our contributions include the curation of a comprehensive dataset of competitive programming problems, standard and erroneous solutions, and the definition of two key evaluation tasks focusing on the generation of valid and targeted test case generators. Crucially, we develop a high-quality dataset for code analysis, demonstrating its utility in enhancing LLM performance through prompting and fine-tuning for generating targeted test cases. TCGBench thus provides a foundational platform for future research to systematically probe and advance LLM capabilities in code verification via robust test case generation.
\section{Limitations}
\label{sec:limitations}

This work is subject to several limitations. 1) Due to the scarcity of competition-level programming (CP) problems, this research utilized a subset of problems with extensive human verification as the dataset to facilitate the investigation; consequently, the difficulty and diversity of the data are restricted. 2) This work focuses on CP problems, while demanding significant reasoning, meaning its findings may not directly translate to software engineering contexts, where code structures differ considerably; establishing a robust benchmark in long-form software engineering scenarios might more accurately reflect real-world development requirements. 3) While this work focuses on building a reliable benchmark for evaluating LLMs in test case generation, it does not explore ways to improve model performance. For example, whether models can improve without human-written target instructions—perhaps via reinforcement learning—remains an open question.


\bibliographystyle{unsrtnat}
\bibliography{ref}

\newpage
\appendix

\section{Detailed Examples and Case Studies}
\label{sec:app_examples}

In this section, we provide detailed examples from our dataset and conduct case studies.

\subsection{Problem Statement}

Figure~\ref{fig:example_full_statement} shows an example of a complete problem statement.

\begin{figure}[H]
  \centering
  \begin{tcolorbox}[blue, title=Problem Description]
A train departs from the starting station (called station $1$) with $a$ passengers boarding at this station. The train then arrives at station $2$, where people get on and off, but the number of people getting on equals the number getting off. Therefore, when departing from station $2$ (i.e., before arriving at station $3$), the number of passengers on the train remains $a$. From station $3$ onwards (including station $3$), the number of people getting on and off follows a certain pattern: the number of people getting on is the sum of the number of people who got on at the previous two stations, while the number of people getting off equals the number of people who got on at the previous station. This pattern continues until the second-to-last station (station $n-1$). Given that there are $n$ stations in total, $a$ people board at the starting station, and $m$ people get off at the last station (all passengers disembark), how many passengers are on the train after leaving station $x$?
  \end{tcolorbox}
  \begin{tcbraster}[
    raster columns=2,
    raster rows=1,
    raster force size=false,
    colback=white!100,
    boxrule=0pt,
    left=0pt, right=0pt, top=0pt, bottom=0pt,
    raster equal height=rows,
    raster column skip=0.01\textwidth
  ]
  \begin{tcolorbox}[coral, title=Input Format,width=0.495\textwidth]
The input consists of only one line with four integers, representing the number of people boarding at the starting station $a$, the number of stations $n$, the number of people getting off at the final station $m$, and the station number $x$ for which the answer is sought.
  \end{tcolorbox}
  \begin{tcolorbox}[coral, title=Output Format, width=0.495\textwidth]
Output a single line with one integer representing the answer: the number of passengers on the train after leaving station $x$.
  \end{tcolorbox}
  \end{tcbraster}
  \begin{tcbraster}[
    raster columns=2,
    raster rows=1,
    raster force size=false,
    colback=white!100,
    boxrule=0pt,
    left=0pt, right=0pt, top=0pt, bottom=0pt,
    raster equal height=rows,
    raster column skip=0.01\textwidth
  ]
  \begin{tcolorbox}[grey, title=Sample Input,width=0.495\textwidth]
\begin{lstlisting}[style=txtSimple]
5 7 32 4
\end{lstlisting}
  \end{tcolorbox}
  \begin{tcolorbox}[grey, title=Sample Output,width=0.495\textwidth]
\begin{lstlisting}[style=txtSimple]
13
\end{lstlisting}
  \end{tcolorbox}
  \end{tcbraster}
  \begin{tcolorbox}[pink, title=Variable Constraints]
For all test cases, it is guaranteed that $1 \leq a \leq 20$, $1 \leq x \leq n \leq 20$, $1 \leq m \leq 2 \times 10^4$.
  \end{tcolorbox}
  \caption{A complete problem statement.}
  \label{fig:example_full_statement}
\end{figure}

\subsection{Erroneous Solver Program}

The following figures illustrate several erroneous solver program examples. Each example includes a problem statement, an erroneous solver program, an example of the corresponding target instruction (which points out errors of the program and how to write a targeted generator), and a targeted test case input that triggers the errors in the code. The problem statements are simplified, and the programs only retain the core algorithm parts for clarity.

\begin{itemize}
  \item Figure \ref{fig:example_error_logic} presents an erroneous solver program example where the algorithm of the code is incorrect.
  \item Figure \ref{fig:example_error_implementation} presents an erroneous solver program example with a correct algorithm but a small implementation error in the code.
  \item Figure \ref{fig:example_error_corner_case} presents an erroneous solver program example that is almost correct but fails on an edge case.
\end{itemize}

\begin{figure}[H]
  \centering
  \begin{tcolorbox}[blue,title=Problem Statement (Simplified)]
    Find two paths on a grid from the top-left corner to the bottom-right corner, where each move can only go either right or down. The goal is to maximize the sum of the numbers covered by the two paths, with overlapping cells only counted once.













  \end{tcolorbox}
  \begin{tcolorbox}[purple, title=Erroneous Solver Program]
    \begin{lstlisting}[style=cppSimple]
int grid[MAXN+1][MAXN+1];
int dp[MAXN+1][MAXN+1];
int max_path(int n, int m) {
  memset(dp, 0, sizeof(dp));
  dp[1][1] = grid[1][1];
  for (int i = 1; i <= n; ++i) {
    for (int j = 1; j <= m; ++j) {
      if (i > 1)
        dp[i][j] = max(dp[i][j], dp[i-1][j] + grid[i][j]);
      if (j > 1)
        dp[i][j] = max(dp[i][j], dp[i][j-1] + grid[i][j]);
    }
  }
  return dp[n][m];
}
int main() {
  int N;
  cin >> N;
  memset(grid, 0, sizeof(grid));
  int x, y, value;

  while (cin >> x >> y) {
    if (x == 0 && y == 0) break;
    cin >> value;
    grid[x][y] = value;
  }
  int first_max = max_path(N, N);
  int grid_copy[MAXN+1][MAXN+1];
  memcpy(grid_copy, grid, sizeof(grid));
  int i = N, j = N;
  while (i > 1 || j > 1) {
    if (i > 1 && dp[i][j] == dp[i-1][j] + grid[i][j]) {
      grid[i][j] = 0;
      i--;
    } else {
      grid[i][j] = 0;
      j--;
    }
  }
  grid[1][1] = 0;
  int second_max = max_path(N, N);
  cout << first_max + second_max << endl;
  return 0;
}
    \end{lstlisting}
  \end{tcolorbox}
  \begin{tcbraster}[
    raster columns=2,
    raster rows=1,
    raster force size=false,
    colback=white!100,
    boxrule=0pt,
    left=0pt, right=0pt, top=0pt, bottom=0pt,
    raster equal height=rows,
    raster column skip=0.01\textwidth
  ]
  \begin{tcolorbox}[orange,title=Target Instruction, width=0.7\textwidth]
    The code first finds a path with the maximum sum, then sets the values of the cells along that path to zero, then finds another path with the maximum sum. This algorithm is incorrect. In the hack case, the first path takes all the grids numbered with $5$, leaving two $1$s that are impossible to collect at once. But there exists a solution that covers the whole grid.
  \end{tcolorbox}
  \begin{tcolorbox}[grey,title=Targeted Test Case Input, width=0.29\textwidth]
    \begin{lstlisting}[style=txtSimple]
3
1 1 5
1 2 1
2 1 5
2 2 5
3 1 1
3 2 5
0 0 0
    \end{lstlisting}
  \end{tcolorbox}
  \end{tcbraster}
  \caption{An erroneous solver program example. The algorithm of the code is incorrect.}
  \label{fig:example_error_logic}
\end{figure}

\begin{figure}[H]
  \centering
  \begin{tcolorbox}[blue,title=Problem Statement (Simplified)]
    Determine whether it is possible to transform a string $a$ into $b$ using a set of transformations within $10$ steps. If possible, output the number of the minimum steps.

    A transformation is defined as a pair of strings $(p,q)$. If $p$ is a substring of $a$, it can be replaced with $q$. If multiple positions of $p$ exist in $a$, you can only replace one in a step.
  \end{tcolorbox}
  \begin{tcolorbox}[purple, title=Erroneous Solver Program]
    \begin{lstlisting}[style=cppSimple]
string a, b;
unordered_map<string, string> transformations;
queue<pair<string, int>> q;
int main() {
  cin >> a >> b;
  string tempA, tempB;
  while (cin >> tempA >> tempB)
    transformations[tempA] = tempB;
  q.push({a, 0});
  while (!q.empty()) {
    pair<string, int> current = q.front();
    q.pop();
    if (current.second >= 10)
      continue;
    for (const auto& [from, to] : transformations) {
      int pos = 0;
      while ((pos = current.first.find(from, pos)) != 
              string::npos) {
        string temp = current.first.substr(0, pos) + 
                      to + 
                      current.first.substr(pos + from.size());
        if (temp == b) {
          cout << current.second + 1;
          return 0;
        } else {
          q.push({temp, current.second + 1});
          pos++;
        }
      }
    }
  }
  cout << "NO ANSWER!" << endl;
  return 0;
}
    \end{lstlisting}
  \end{tcolorbox}
  \begin{tcbraster}[
    raster columns=2,
    raster rows=1,
    raster force size=false,
    colback=white!100,
    boxrule=0pt,
    left=0pt, right=0pt, top=0pt, bottom=0pt,
    raster equal height=rows,
    raster column skip=0.01\textwidth
  ]
  \begin{tcolorbox}[orange,title=Target Instruction, width=0.7\textwidth]
    The code uses an unordered\_map to store the transformations, which is incorrect since there might be multiple transformations for the same string. The hack case gives such a set of transformations. The code will only store the last transformation, making it impossible to transform $a$ into $b$. But it is possible using the second transformation.
  \end{tcolorbox}
  \begin{tcolorbox}[grey,title=Targeted Test Case Input, width=0.29\textwidth]
    \begin{lstlisting}[style=txtSimple]
abc
axy

bc pq
bc xy
bc st
    \end{lstlisting}
  \end{tcolorbox}
  \end{tcbraster}
  \caption{An erroneous solver program example. There is a small implementation error in the code.}
  \label{fig:example_error_implementation}
\end{figure}

\begin{figure}[H]
  \caption{An erroneous solver program example. The code fails on an edge case.}
  \label{fig:example_error_corner_case}
  \centering
  \begin{tcolorbox}[blue,title=Problem Statement (Simplified)]
    Calculate the value of $a_1+a_2+\ldots+a_n$, where $a=\{1,2,2,3,3,3,4,4,4,4,\ldots\}$.
  \end{tcolorbox}
  \begin{tcolorbox}[purple, title=Erroneous Solver Program]
    \begin{lstlisting}[style=cppSimple]
int main(){
  int n;
  int s=0,m;
  cin>>n;
  for(int i=1;i<n;i++){
    s=s+i;
    if(s>=n){
      m=i;
      break;
    }
  }
  int S=0;
  for(int i=1;i<m;i++){
    S=S+i*i;
  }
  S=S+(n-s+m)*m;
  cout<<S;
}
    \end{lstlisting}
  \end{tcolorbox}
  \begin{tcbraster}[
    raster columns=2,
    raster rows=1,
    raster force size=false,
    colback=white!100,
    boxrule=0pt,
    left=0pt, right=0pt, top=0pt, bottom=0pt,
    raster equal height=rows,
    raster column skip=0.01\textwidth
  ]
  \begin{tcolorbox}[orange,title=Target Instruction, width=0.7\textwidth]
    The code failed to handle the edge case when $n=1$. In this case, the first loop will not run, and $m$ will be uninitialized. So the code will finally output a garbage value.
  \end{tcolorbox}
  \begin{tcolorbox}[grey,title=Targeted Test Case Input, width=0.29\textwidth]
    \begin{lstlisting}[style=txtSimple]
1
    \end{lstlisting}
  \end{tcolorbox}
  \end{tcbraster}
\end{figure}

\newpage

\subsection{Case Study of Generating Test Case Generators}

Figure~\ref{fig:full_example_benchmark} illustrates a full example of generating a targeted generator. The example includes a problem statement, a valid generator, an erroneous solver program, a correct target instruction for reference, and a targeted generator that triggers the error in the code.


\begin{center}
  \centering
  \begin{tcolorbox}[blue,title=Problem Statement (Simplified)]
    Implement an Aho-Corasick automaton. Given a text string $S$ and $n$ pattern strings $T_{1 \sim n}$. For each pattern string $T_i$, determine how many times it appears in $S$.

    For $100\%$ of the data, $1 \le n \le 2 \times {10}^5$, the total length of $T_{1 \sim n}$ does not exceed $2 \times {10}^5$, and the length of $S$ does not exceed $2 \times {10}^6$.
  \end{tcolorbox}
  \begin{tcolorbox}[cyan, title=Valid Test Case Generator]
    \begin{lstlisting}[style=pythonSimple]
import random
import string

def generate_string(length):
    return ''.join(random.choice(string.ascii_lowercase) for _ in range(length))

def construct_inputs():
    n = 200000
    total_length_T = random.randint(1, 2 * 10**5)  # total length of all pattern strings <= 200000
    S_length = random.randint(1, 2 * 10**6)  # length of S <= 2000000

    # Generate pattern strings
    patterns = []
    current_length = 0
    while len(patterns) < n and current_length < total_length_T:
        # Random length for each pattern string
        pattern_length = random.randint(1, min(20, total_length_T - current_length))  # limit to 20 for practicality
        pattern = generate_string(pattern_length)
        patterns.append(pattern)
        current_length += pattern_length

    # Generate the main string S
    S = generate_string(S_length)

    inputs = [str(n)] + patterns + [S]
    return "\n".join(inputs)

if __name__ == "__main__":
    generated_output = construct_inputs()
    print(generated_output)
    \end{lstlisting}
  \end{tcolorbox}
  \begin{tcolorbox}[breakable, purple, title=Erroneous Solver Program]
    \begin{lstlisting}[style=cppSimple]
#include <bits/stdc++.h>
using namespace std;
using ll = long long;
const int N = 5e6 + 10;
struct AhoCorasick
{
    static constexpr int ALPHABET = 26;
    struct Node {
        int len;
        int link;
        std::array<int, ALPHABET> next;
        Node() : len{0}, link{0}, next{} {}
    };

    std::vector<Node> t;
    AhoCorasick(){
        init();
    }

    void init() {
        t.assign(2, Node());
        t[0].next.fill(1);
        t[0].len = -1;
    }

    int newNode() {
        t.emplace_back();
        return t.size() - 1;
    }

    int add(const std::string &a) {
        int p = 1;
        for (auto c : a) {
            int x = c - 'a';
            if (t[p].next[x] == 0) {
                int k = newNode();
                t[p].next[x] = k;
                t[t[p].next[x]].len = t[p].len + 1;
            }
            p = t[p].next[x];
        }
        return p;
    }

    void work() {
        std::queue<int> q;
        q.push(1);

        while (!q.empty()) {
            int x = q.front();
            q.pop();

            for (int i = 0; i < ALPHABET; i++) {
                if (t[x].next[i] == 0) {
                    t[x].next[i] = t[t[x].link].next[i];
                } else {
                    t[t[x].next[i]].link = t[t[x].link].next[i];
                    q.push(t[x].next[i]);
                }
            }
        }
    }

    int next(int p, int x) {
        return t[p].next[x];
    }

    int link(int p) {
        return t[p].link;
    }

    int len(int p) {
        return t[p].len;
    }

    int size() {
        return t.size();
    }
};

ll n, ans[N];
AhoCorasick a;
int main()
{
    ios::sync_with_stdio(false);
    cin.tie(0);
    cin >> n;
    vector<int>vx;
    string ss;
    for (int i = 1; i <= n; i++)
    {
        string s;
        cin >> s;    
        int k = a.add(s);
        vx.push_back(k);
    }
    cin >> ss;
    a.work();
    int st = 1;
    for (auto c: ss)
    {
        st = a.t[st].next[c - 'a'];
        ans[st]++;
    }
    for (int i = a.size() - 1; i >= 1; i--) 
    {
        auto u = a.t[i];
        ans[u.link] += ans[i];
    }
    for (auto v: vx)
        cout << ans[v] <<"\n";
}
    \end{lstlisting}
  \end{tcolorbox}
  \begin{tcolorbox}[orange,title=Code Issue Document]
The core issue in the implementation stems from incorrectly initializing the root node's transition table to point to node 1, rather than 0. In Aho-Corasick automaton, failed transitions should fall back to the root (node 0), ensuring that unmatched characters do not lead to undefined behavior. Redirecting all root transitions to node 1 disrupts the construction of failure links and introduces incorrect fallback paths, especially under complete mismatches. This misconfiguration compromises the integrity of the automaton and leads to erroneous pattern matching counts.
  \end{tcolorbox}
  \begin{tcolorbox}[brown,title=Hands-on Instruction]
To generate a test case that make the code fail, we need to create input where:

1. Use pattern strings where the short patterns are the suffix or prefix of the longer ones. (e.g. \texttt{a},\texttt{ab},\texttt{cab},\texttt{abc})

2. The text string contains repeated overlaps of these patterns.

Follow these steps:

1. Use multiple pattern strings that are easy to overlap.

2. Generate a text string that contains multiple repeating or overlapping pattern strings.
  \end{tcolorbox}
  \begin{tcolorbox}[green,title=Successful Targeted Generator]
    \begin{lstlisting}[style=pythonSimple]
import random
import string

def construct_inputs():
    # Max constraints
    n = random.randint(1, 200000)  # Random number of patterns up to 200000
    patterns = []
    
    # Generate patterns with overlapping occurrences
    for _ in range(n):
        length = random.randint(1, 10)  # Pattern length ranging from 1 to 10
        pattern = ''.join(random.choices(string.ascii_lowercase, k=length))
        patterns.append(pattern)

    # Generate the main string S that includes many occurrences
    S_length = 2000000  # Close to the limit
    # Including a concat of patterns to ensure they appear multiple times
    S = ''.join(random.choices(patterns, k=S_length // 10))  # Create many copies of patterns

    inputs = []
    inputs.append(str(n))
    inputs.extend(patterns)
    inputs.append(S)

    return "\n".join(inputs)

def main():
    print(construct_inputs())
    
if __name__ == "__main__":
    main()
    \end{lstlisting}
  \end{tcolorbox}
  \captionof{figure}{A successful example of generating a valid generator and a targeted generator.}
  \label{fig:full_example_benchmark}
\end{center}

We next analyze the effect of target instruction. As illustrated in Figure~\ref{fig:without_hack_instruction} and Figure~\ref{fig:with_hack_instruction}, the model tends to hallucinate in the absence of a target instruction, often resulting in failing to identify the error. The hallucinatory tendency is significantly mitigated when we provide the target instruction.


\begin{center}
  \begin{tcolorbox}[breakable,blue, title=Problem Statement (Simplified)]
    Given a decimal number $n$ and a base $r$, convert the decimal number to the base $r$ representation. The base $r$ can be negative.
  \end{tcolorbox}
  \begin{tcolorbox}[purple, title=Erroneous Solver Program]
    \begin{lstlisting}[style=cppSimple]
#include <bits/stdc++.h>
using namespace std;
void zhuan(int n,int r) {
    if(n==0) return;
    int m=n%r;
    if (m<0) m-=r,n+=r;
    if (m>=10) m='A'+m-10;
    zhuan(n/r,r);
    cout<<m;
}
int main() {
    int n,r;
    string ans="";
    cin>>n>>r;
    cout<<n<<"=";
    zhuan(n,r);
    printf("(base%d)",r);
    return 0;
}
    \end{lstlisting}
  \end{tcolorbox}
  \begin{tcolorbox}[red, title=An Unsuccessful Targeted Generator Generated By LLM]
    \begin{lstlisting}[style=pythonSimple]
import sys

def construct_inputs():
    # This input will cause the given C++ program to produce incorrect output
    # due to mishandling of negative remainders in recursion
    n = -1
    r = -2
    test_input = f"{n} {r}"
    return test_input

def main():
    # Output the test input
    sys.stdout.write(construct_inputs())

if __name__ == "__main__":
    main()
    \end{lstlisting}
  \end{tcolorbox}
  \captionof{figure}{An example of LLM failing to generate a successful targeted generator when we don't provide a target instruction.} \label{fig:without_hack_instruction}
\end{center}

\newpage

\begin{center}
  \begin{tcolorbox}[breakable, orange, title = Code Issue Document]
The code has an error in the digit output logic. According to the problem requirements, when converting a number to a negative base representation, digits greater than 9 need to be represented as letters (A, B, C, etc.) in the output.

The code correctly identifies digits greater than or equal to 10 in this line:

\begin{lstlisting}[style=cppSimple]
if (m>=10) m='A'+m-10;
\end{lstlisting}

However, it incorrectly outputs the digit directly:

\begin{lstlisting}[style=cppSimple]
cout<<m;
\end{lstlisting}

This is incorrect because:
1. When `m` is converted to a character value (like 'A', 'B', etc.), it becomes a char type.
2. When outputting the char directly without type conversion, it will print the ASCII character instead of the digit/letter representation.
3. For example, if m='A', it will print the actual character 'A' instead of the digit value 10.

The correct approach would be to explicitly convert the value to char when outputting, or use separate output logic for numeric and letter values. 
  \end{tcolorbox}
  
  \begin{tcolorbox}[breakable, brown, title=Hands-on Instruction]
To generate a test case that will make the code fail, we need to create input where:

1. The base is negative with an absolute value greater than 10 (e.g., -16);

2. When converting the number to this negative base, at least one digit would need to be represented by a letter (value $\ge$ 10);

Follow these steps:

1. Choose a negative base with absolute value > 10, such as -16;

2. Generate a decimal number that, when converted to the chosen negative base, will have at least one digit with value $\ge$ 10;

3. Since the erroneous code incorrectly outputs character values, it will display incorrect ASCII characters instead of the proper digits;

The targeted generator should:

1. Randomly select a negative base between -20 and -11 (with abs value > 10);

2. Generate a suitable decimal number that would require letter representation in that base;

3. Ensure the absolute value of the number is within the problem constraint (|n| $\le$ 37336);

4. The test case should produce output where instead of seeing letters A-F for digits 10-15, you'll see incorrect ASCII characters. 
  \end{tcolorbox}
  \begin{tcolorbox}[breakable, green, title=A Successful Targeted Generator Generated By LLM]
    \begin{lstlisting}[style=pythonSimple]
import random
import math

# We simulate the conversion process as done in the given C++ code.
def convert(n, r):
    # Special case: if n is zero, the proper conversion is "0"
    if n == 0:
        return "0"
    digits = []
    num = n
    while num != 0:
        original = num
        # Use truncation toward zero (like C++ division)
        q = math.trunc(num / r)
        m = num - q * r
        if m < 0:
            m = m - r
            num = original + r
        else:
            num = q
        # If m is 10 or more, then it should be represented as a letter
        if m >= 10:
            # Convert to letter ('A' for 10, 'B' for 11, etc.)
            digits.append(chr(ord('A') + m - 10))
        else:
            digits.append(str(m))
    # The conversion uses recursion so that the leftmost digit is computed last,
    # hence we need to reverse our digit list.
    return "".join(reversed(digits))

def construct_inputs():
    # We need to choose a negative base with an absolute value > 10.
    # So pick randomly between -20 and -11.
    while True:
        r = -random.randint(11, 20)
        # We try several candidate n values (both positive and negative) within |n| <= 37336.
        candidates = [-37336, -30000, -25000, -20000, 20000, 25000, 30000, 37336]
        random.shuffle(candidates)
        for n in candidates:
            rep = convert(n, r)
            # Check if any character in the representation is an alphabet letter.
            # That is, a digit that should be printed as a letter.
            if any(ch.isalpha() for ch in rep):
                # Return the input with n and r.
                # Input format: each line of input contains two numbers.
                # We output a single test case.
                return f"{n} {r}"
    # In the unlikely event no candidate qualifies, default to a known hack case.
    return "-25000 -16"

def main():
    print(construct_inputs())

if __name__ == "__main__":
    main()
    \end{lstlisting}
  \end{tcolorbox}
  \captionof{figure}{The corresponding example of LLM to generate a successful targeted generator when we provide a target instruction. The problem statement and the erroneous solver program are the same as Figure~\ref{fig:without_hack_instruction}.} 
  \label{fig:with_hack_instruction}
\end{center}

\subsection{Details of Fine-Tuning}


As outlined in Section 4, we conduct supervised fine-tuning (SFT) of our model utilizing two NVIDIA RTX A6000 GPUs with 48GB memory capacity each. All training samples are formatted using the Alpaca-style instruction template listed below. As our training set is sourced entirely from the NOIP Dataset, we will subsequently use the Canonical Dataset for evaluation. This serves the dual purpose of ensuring no data overlap and enabling a robust evaluation of the model's generalization performance, particularly because the problem types within these two datasets are significantly different.

\begin{figure}[H]
  \centering
  \begin{tcolorbox}[blue,title=Alpaca-style instruction template]
    \begin{lstlisting}
Below is an instruction that describes a task, paired with an input that provides further context. Write a response that appropriately completes the request.

### Instruction:
{ An instructional system prompt }

### Input:
{ A problem statement and an erroneous solver program }

### Response:
{ Target instructions and the corresponding targeted generator }
    \end{lstlisting}
  \end{tcolorbox}
  \caption{An example of the Alpaca-style instruction template.}
  \label{fig:example_instruction}
\end{figure}

After fine-tuning, we test its performance on the canonical dataset. We find a significant enhancement (refer to Section \ref{sec:results}) between the original model and the fine-tuned one. We present our case studies in Figure~\ref{fig:example_vanilla_qwen} and Figure~\ref{fig:example_finetuned_qwen}. The result suggests that the model can learn common error patterns and a better generator construction technique from the supervised data.

\begin{center}
  \begin{tcolorbox}[breakable, blue, title=Problem Statement (Simplified)]
    Given a system of linear equations, solve it.
    $$
    \begin{cases}
    a_{1, 1} x_1 + a_{1, 2} x_2 + \cdots + a_{1, n} x_n = b_1 \\
    a_{2, 1} x_1 + a_{2, 2} x_2 + \cdots + a_{2, n} x_n = b_2 \\
    \cdots \\
    a_{n,1} x_1 + a_{n, 2} x_2 + \cdots + a_{n, n} x_n = b_n
    \end{cases}
    $$
  \end{tcolorbox}
  \begin{tcolorbox}[breakable, purple, title=Erroneous Solver Program]
    \begin{lstlisting}[style=cppSimple]
#include <bits/stdc++.h>
using namespace std;
#define int  long long
// typedef long long ll;
#define inf 0x3f3f3f3f
#define fo(i,l,r) for(int i=l;i<=r;i++)
#define foo(i,l,r) for(int i=l;i>=r;i--)
#define yes cout<<"yes\n"
#define no cout<<"no\n"
#define YES cout<<"YES\n"
#define NO cout<<"NO\n"
#define Yes cout<<"Yes\n"
#define No cout<<"No\n"
#define sl s<<1
#define sr s<<1|1
#define m(l,r) l+r>>1
#define pp pair<int,int>
#define ppp pair<pair<int,int>,int>
#define int128 __int128
const int mod=998244353;
int lowbit(int x){return x&-x;}
int qsm(int l,int r){int ans=1;while(r){if(r&1)ans=ans*l%mod;l=l*l%mod;r>>=1;}return ans;}
const int e=5e5+100;
const int ee=1e6+10;
int a,b,l,r,v,c,t,k;
int cnt=1,cnt2=1;
int pd[e];
int d[e];
// int pd2[e];
struct me{int l,r,v;}m[e<<1];
void add(int l,int r,int v=0){m[cnt]={r,pd[l],v};pd[l]=cnt++;}
// void add2(int l,int r,int v=0){mm[cnt2]={r,pd2[l],v};pd2[l]=cnt2++;}
string ans;
int x,y,z=1;
int s=0,ss=0,top=0,sss=0;
bool vis[e];
double dp[60][60];
void again(){
cin>>a;
fo(i,1,a)fo(j,1,a+1)cin>>dp[i][j];
fo(i,1,a){
    s=i;
    fo(j,i+1,a)if(fabs(dp[j][i])>fabs(dp[s][i]))s=j;
    fo(j,1,a+1)swap(dp[s][j],dp[i][j]);
    if(fabs(dp[i][i])<1e-7)return puts("No Solution"),void();
    foo(j,a+1,1)dp[i][j]/=dp[i][i];
    fo(j,1,a){
        if(j==i)continue;
        double v=dp[j][i]/dp[i][i];
        fo(k,1,a+1)dp[j][k]-=v*dp[i][k];
    }
}
fo(i,1,a)cout<<fixed<<setprecision(2)<<dp[i][a+1]<<"\n";
}

signed main(){
ios::sync_with_stdio(false);cin.tie(0);cout.tie(0);
t=1;
while(t--)again();
return 0;
}
    \end{lstlisting}
  \end{tcolorbox}
  \begin{tcolorbox}[breakable, red, title=The Original Model's Response]
    \begin{lstlisting}
## Code Issues

The provided code attempts to solve the problem of finding the longest common subsequence of two sequences. However, the code has't consider the following conditions correctly:
- The code fails to handle the case where the input sequence is large enough. This causes the code to fail to handle the case where the input sequence is large.
- The code doesn't handle the case where the input sequence is long enough. This cause the code to fail to handle the case where the input sequence is long enough.
- The code doesn't handle the case where the input sequence is short enough.
## Code Issues

The main issue with the provided code is that it doesn doesn't handle the case where the input sequence is short enough. This means the code won't work properly when the input sequence is too long. This can be seen in the case where the input sequence is is too short.
The code fails to handle the case where the input sequence is long enough. This can be seen in the case where the input sequence is too short. The code fails to handle the case where the input sequence is too long. This can be seen in the case where the input sequence is too short.
The code fail to handle the case where the input is too long. This can be seen in the case where the input is too short.
The code doesn't handle the case where the input is too long. This can be seen in the case where the input is too short.
The code fail to manage the scenario where the sequence is too short. This can be seen in the situation where the input is too long.
The code doesn't address the situation where the input is too short. This can be seen in the instance where the input is too long.
```python
import random

from typing import List
import string
import sys
import random

def construct_inputs():
    # Set w to a value between 80 and 200
    w = random.randint(80, 200)
    n = random.randint(3times, 3 * 10**4)
    return n
    prices = [random.randint(5, w) for _ in range(n)]
    inputs = []
    inputs_sequence = []
    for i in range(n):
        inputs.append(random.randint(5, w))
    inputs = []
    inputs = []
    for price in prices:
       inputs.append(str(i) for cases in inputs)
    return inputs
print(construct_inputs())
return inputs
def main():
    print(main())
    inputs = []
    for price in prices:
        inputs.append(str(i))
for i in range(n):
for i in range(n):
    input = random.choices(string.ascii_letters for _ in range(n))
    for i in range(n):
       input.append(str(i) for price in prices)
    for i in range(n):
       input.append(str(i)) j(range(n))
    for price in prices:
    inputs.append(str(pi) for pi in prices)
    for pi in prices:
       inputs.append(string.print(fprices[i]) for price in prices)
    return input_sequence
print(construct_inputs()
return input
```
## Hack Instruction

To generate a test case that exposes the flaw in the program, the input should have an odd number of elements and the sum of the prices in each group must exceed the upper bound. This ensures the program fails to handle the case where there are an odd number of elements and the sum of prices in each group does not exceed the upper bound for the sum of prices in each group.
To see the problem statement and the input format, the code should have the input of the number of test cases set to 1 and the number of souvener.ApplyResources()
    \end{lstlisting}
  \end{tcolorbox}
  \captionof{figure}{An example of the original model's response. In this example, the original model demonstrates a significant tendency towards hallucination, frequently exhibiting repetitive textual patterns and an inability to produce a targeted (or even valid) generator. Such deficiencies are common throughout the responses generated by the original model.}
  \label{fig:example_vanilla_qwen}
\end{center}

\begin{center}
  \captionof{figure}{The corresponding example of the fine-tuned model's response. It can be observed that the output quality of the model is significantly improved.}
  \label{fig:example_finetuned_qwen}
  \centering
  \begin{tcolorbox}[breakable, green, title=The Fine-tuned Model's Response]
    \begin{lstlisting}
## Code Issues

The code attempts to solve a system of linear equations using Gaussian elimination. However, it contains a fundamental error in handling degenerate cases. The problem states that if the system has either no solution or infinitely many solutions, the output should be "No Solution".

The error lies in the handling of the augmented coefficient matrix. After applying row operations, the code checks if a diagonal element is zero (`ez(a[c][x])`) and skips the current variable if it is. This allows the algorithm to potentially produce an incorrect result when the system is under-constrained or inconsistent.

Additionally, the code fails to properly check for inconsistencies or degeneracy after back-substitution. Specifically, in the final loop where it prints the results, it does not verify whether the remaining rows represent valid equations. If the remaining rows contain only zeros in the coefficients but non-zero values in the constant term, this indicates an inconsistency (no solution), but the code doesn't handle this case appropriately.

In summary, the code lacks proper checks for:
- Whether the system is under-constrained or has free variables
- Whether the final equations are consistent after back-substitution

These issues lead to incorrect outputs when the system has no unique solution or no solution at all.

## Hack Instruction

To generate a test case that exposes the flaws in the code, we need to create a system of linear equations that has either:
- No solution (inconsistent system)
- Infinitely many solutions (underdetermined system)

We'll focus on creating an inconsistent system, where the Gaussian elimination process will reveal a row of the form `[0 0 ... 0 | non-zero]`, indicating an impossible equation like `0 = k` (where `k != 0`).

Our strategy will include:
1. Setting up a small system (e.g., 3x3) for easy verification.
2. Creating equations such that after row reduction, a row of zeros in the coefficients but a non-zero constant appears.
3. Ensuring the system conforms to the problem constraints: `1 <= n <= 100`, `|a_i| <= 10^4`, and `|b| <= 10^4`.

By doing this, we can ensure the code outputs an incorrect result, as it lacks proper checks for inconsistent systems.
```python
import random

def construct_inputs():
    """
    Construct a 3x3 system of linear equations that is inconsistent.
    This means we'll create equations such that after Gaussian elimination,
    we get a row of the form [0 0 ... 0 | non-zero], indicating an impossible equation.
    """
    # We'll create a simple 3x3 system
    n = 3
    
    # Coefficients and constants for our equations
    # We'll manually create an inconsistent system
    # Equations: 
    #   x +  y     - z =  2
    #   2x + 2y - 2z =  4 (multiple of the first equation)
    #   x +  y     - z =  3 (same as the first but with different constant)
    # After elimination, this will produce a row like [0 0 0 | 1], meaning 0 = 1
    
    # Manually specify the coefficients and constants
    # Format: a[3][4] where a[i][j] represents a_ij in the equation
    a = [
        [1, 1, -1, 2],  # First equation: x + y - z = 2
        [2, 2, -2, 4],  # Second equation: 2x + 2y - 2z = 4 (same as first * 2)
        [1, 1, -1, 3]   # Third equation: x + y - z = 3 (same as first but with different constant)
    ]
    
    # Convert to string format
    inputs = []
    inputs.append(str(n))
    for i in range(n):
        row_str = ' '.join(map(str, a[i]))  # Convert each row to a space-separated string
        inputs.append(row_str)
    
    return '\n'.join(inputs)

def main():
    print(construct_inputs())

if __name__ == "__main__":
    main()
```
    \end{lstlisting}
  \end{tcolorbox}

\end{center}

\newpage
\section{Prompts}

Figure~\ref{fig:prompt_basic} illustrates the prompt for generating a valid test case generator. Figure~\ref{fig:prompt_hack_without_instructions} and Figure~\ref{fig:prompt_hack_with_instructions} illustrate the prompt for generating a targeted test case generator without target instruction and the prompt with target instruction, respectively. 

\begin{figure}[H]
  \centering
  \begin{tcolorbox}[olive, title=Prompt of Generating a Valid Test Case Generator]
    \begin{lstlisting}
You are an assistant in a competitive programming contest. You should construct input-generators for testing programming contest problems. Your generator should be a python program, with a function 'construct_inputs' that returns an input test case sampled from the generator. To avoid unnecessary errors, please pay attention to the instructions and constraints present in the problem statement and input formats.

Your method should be based on random algorithms, instead of hard-coding the inputs.

The detailed rules are as follows:
## Generator Coding
- Give your python3 input-generators. Other languages are considered invalid.
- `construct_inputs` and `main` are required. It is allowed to use other functions or anything you want to help construct inputs.
- Give only one complete program instead of dividing it into multiple parts.
- Some values of the input variables will be provided. If the values are given, you need to strictly follow these values and write them in `construct_inputs`. Other variables should be generated randomly with the given constraints in the problem statement.
- Wrap your code in markdown ```python ... ```.
## Prohibited Actions
- Use special characters that cannot be decoded by UTF-8 or do not conform to general specifications.

Notices: 
Please strictly follow the input in the problem formats. Do not imagine the input formats, especially whether a problem has multiple test cases. 
NOTE again that if the input includes 'the number of test cases', you should let it be 1 in your generators.

Below is the corresponding valid generator:

{EXAMPLE GENERATOR CODE}
    \end{lstlisting}
  \end{tcolorbox}
  \caption{The prompt template for generating a valid test case generator.}
  \label{fig:prompt_basic}
\end{figure}

\newpage

\begin{figure}[H]
  \centering
    \begin{tcolorbox}[olive, title=Prompt for Generating a Targeted Test Case Generator]
        \begin{lstlisting}
You are a master hacker assistant in a competitive programming contest. You should find errors in user's program and write a hack test case generator. Meanwhile, to avoid unnecessary errors, please pay attention to the instructions and constraints present in the problem statement and input formats.

The detailed rules are as follows:
## Generator Coding
- Give your python3 input-generators. Other languages are considered invalid.
- `construct_inputs` and `main` is required. It is allowed to use other functions or anything you want to help construct inputs.
- Give only one complete program instead of dividing it into multiple parts.
- Your inputs should strictly follow the instructions and constraints present in the problem statement and input formats.
- Wrap your code in markdown ```python ... ```.
- Please NOTE that if the input includes "the number of test cases", you should let it be 1 in your generators.
## Prohibited Actions
- Use special characters which cannot be decoded by UTF-8 or do not conform to general specifications.

Now, you will be given a problem statement and a cpp program that is used to solve the problem, but it contains flaws. First, you need to output the reason why it is wrong. Second, You need to generate a targeted test case generator, ensuring that the generated data can expose flaws of that program related to time complexity issues, logic errors, or overlooked corner cases. Please state the issues and hack instructions, then show the code of your generator.

{EXAMPLE CODE ISSUES}

Here is the example of the problem:
{EXAMPLE STATEMENT}

Here is the wrong example program:
{EXAMPLE CODE}

{EXAMPLE HACK GENERATOR CODE}
        \end{lstlisting}
    \end{tcolorbox} 
  \caption{The prompt for generating a targeted test case generator without target instruction.}
  \label{fig:prompt_hack_without_instructions}
\end{figure}

\begin{figure}[H]
  \centering
    \begin{tcolorbox}[olive, title=Prompt for Generating a Targeted Test Case Generator with Target Instruction]
        \begin{lstlisting}
You are a master hacker assistant in a competitive programming contest. You should find errors in user's program and write a hack test case generator. Meanwhile, to avoid unnecessary errors, please pay attention to the instructions and constraints present in the problem statement and input formats.

The detailed rules are as follows:
## Generator Coding
- Give your python3 input-generators. Other languages are considered invalid.
- `construct_inputs` and `main` is required. It is allowed to use other functions or anything you want to help construct inputs.
- Give only one complete program instead of dividing it into multiple parts.
- Your inputs should strictly follow the instructions and constraints present in the problem statement and input formats.
- Wrap your code in markdown ```python ... ```.
- Please NOTE that if the input includes "the number of test cases", you should let it be 1 in your generators.
## Prohibited Actions
- Use special characters which cannot be decoded by UTF-8 or do not conform to general specifications.

Now, you will be given a problem statement and a cpp program that is used to solve the problem, but it contains flaws. The issue of this program and an instruction of how to hack this code are also given. You need to follow them to generate a targeted test case generator, ensuring that the generated data can expose flaws of that program related to time complexity issues, logic errors, or overlooked corner cases. Please state how you want to hack first, then show the code of your generator.

{EXAMPLE CODE ISSUES}

Here is the example of the problem:
{EXAMPLE STATEMENT}

Here is the wrong example program:
{EXAMPLE CODE}

Here is the example of the issues and hack instructions:
{EXAMPLE CODE ISSUES}

{EXAMPLE HACK GENERATOR CODE}
        \end{lstlisting}
    \end{tcolorbox}
  \caption{The prompt for generating a targeted test case generator with target instruction.}
  \label{fig:prompt_hack_with_instructions}
\end{figure}

\newpage
\section{Details of Target Instruction  Dataset}
\label{sec:app_ti_dataset}

\subsection{Data Format}

   In our target instruction dataset, each data sample is represented as a \texttt{(prompt, label)} pair, where:

\begin{itemize}
  \item \texttt{prompt}: The prompt same as our second task (detailed in Section~\ref{subsec:task_target}), the generation of the targeted test case generator.
  \item \texttt{label}: The corresponding target instruction and the targeted generator. 
\end{itemize}

Specifically, a target instruction consists of a code issue document and a hands-on instruction.






\subsection{Code Issue Document}


The code issue document should describe the location of the error in the code, following a standard template. An example is shown in Figure \ref{fig:example_template_code_issues}.

\begin{figure}[H]
    \centering
\begin{tcolorbox}[orange, title=Code Issue Document]
The problem statement requires ... However, the implementation ... This results in an error when ...
\end{tcolorbox}
    \caption{Code Issue Document Example Template}
    \label{fig:example_template_code_issues}
\end{figure}

\subsection{Hands-On Instruction}


The hands-on instruction specifies how to write the targeted generator. An example template is provided in Figure~\ref{fig:example_template_hack_instruction}.

\begin{figure}[H]
    \centering
\begin{tcolorbox}[brown, title=Hands-on Instruction]
To generate a test case that causes the code to fail, we construct a generator with the following steps:

1. ...

2. ...
\end{tcolorbox}
    \caption{Hands-On Instruction Example Template}
    \label{fig:example_template_hack_instruction}
\end{figure}

The steps detail the precise method for writing the targeted generator.


\subsection{Example Targeted Generator}


An example targeted generator guarantees that
\begin{itemize}
  \item achieving a 100\% error-triggering rate,
  \item strictly following the input format required by the problem statement, and
  \item including complete comments explaining each step.
\end{itemize}

An example template is provided in Figure \ref{fig:example_template_gen}.

\begin{figure}[H]
    \centering
    \begin{tcolorbox}[green, title=Targeted Generator]
    \begin{lstlisting}[style=pythonSimple, numbers=none]
import random

def construct_inputs():
    // Some explanations about the generator
    n = ...
    for i in range(n):
        // Some explanations about the generator
        ...

    inputs = []
    ...

    return "\n".join(inputs)

def main():
    print(construct_inputs())

if __name__ == "__main__":
    main()
    \end{lstlisting}
    \end{tcolorbox}
    \caption{Targeted Generator Example Template}
    \label{fig:example_template_gen}
\end{figure}

\subsection{Full Example}

A full target instruction example is provided in Figure~\ref{fig:full_example_hack}.

\begin{figure}
  \centering
  \begin{tcolorbox}[blue,title=Problem Statement (Simplified)]
    A date is represented as an 8-digit integer in the format YYYYMMDD. Your task is to calculate the number of days whose date representation is a palindrome between two given dates. A date representation is a palindrome if it reads the same backward as forward. For example, January 2, 2010 is represented as 20100102, which is a palindrome.
  \end{tcolorbox}
  \begin{tcolorbox}[purple, title=Erroneous Solver Program]
    \begin{lstlisting}[style=cppSimple]
int a[13] = { 0,31,28,31,30,31,30,31,31,30,31,30,31 };

for (int i = 1; i <=12; i++) {
  for (int j = 1; j <=a[i]; j++) {
    int x = j % 10 * 1000 + j / 10 * 100 + i % 10 * 10 + i/10;
    int z = x * 10000 + i * 100 + j;
    if (z<n || z>m)continue;
    ans++;
  }
}
    \end{lstlisting}
  \end{tcolorbox}
  \begin{tcbraster}[
    raster columns=2,
    raster rows=1,
    raster force size=false,
    colback=white!100,
    boxrule=0pt,
    left=0pt, right=0pt, top=0pt, bottom=0pt,
    raster equal height=rows,
    raster column skip=0.01\textwidth
  ]
  \begin{tcolorbox}[orange,title=Code Issue Document, width=0.5\textwidth]
The problem states that February has 29 days in a leap year and 28 days in a normal year. However, the code uses a fixed array a[13] with February always set to 28 days.

The code never checks whether the current year is a leap year, so it always assumes February has 28 days. This means the code will miss valid palindrome dates that occur on February 29th in leap years.
  \end{tcolorbox}
  \begin{tcolorbox}[brown,title=Hands-on Instruction, width=0.49\textwidth]

To generate a test case that the code will fail, we need to create a generator that:

1. Find a palindrome date that falls on February 29th in a leap year.
2. Makes sure this date is a valid 8-digit palindrome (ABCDDCBA format).
3. Constructs an input with this date as both the starting and ending date.
  \end{tcolorbox}
  \end{tcbraster}
  \begin{tcolorbox}[green, title=Targeted Generator]
    \begin{lstlisting}[style=pythonSimple]
def is_leap_year(year):
    return (year % 4 == 0 and year % 100 != 0) or (year % 400 == 0)

def is_palindrome(date_str):
    return date_str == date_str[::-1]

def construct_inputs():
    year = 9220
    assert is_leap_year(year)
    date_str = f"{year}0229"
    assert is_palindrome(date_str)
    return f"{date_str}\n{date_str}"

def main():
    print(construct_inputs())
    
if __name__ == "__main__":
    main()
    \end{lstlisting}
  \end{tcolorbox}
  \caption{A full example of a target instruction data sample.}
  \label{fig:full_example_hack}
\end{figure}








\newpage
\section{Tabular Results}
\label{sec:app_tabular_results}

We now provide the full experimental results in Table~\ref{tab:basic}, Table~\ref{tab:hack}, and Table~\ref{tab:noip_with_instrns}.

\begin{table}[h]
  \centering
  \caption{Valid@k results of generating valid generators in two datasets, respectively.}
  \vspace{1ex}
  \label{tab:basic}
  \begin{tabular}{@{}l|cccc@{}}
    \toprule
    \multicolumn{5}{l}{\textbf{NOIP Dataset}} \\ 
    \midrule
    Model Name       & Valid @1 & Valid @3 & Valid @5 & Valid @10 \\
    \midrule
    GPT-4o-mini      & $0.689$ & $0.761$ & $0.781$ & $0.802$ \\
    GPT-4o           & $0.765$ & $0.815$ & $0.825$ & $0.832$ \\
    GPT-4            & $0.868$ & $0.918$ & $0.925$ & $0.931$ \\
    GPT-3.5-Turbo    & $0.696$ & $0.863$ & $0.893$ & $0.911$ \\
    o1-mini          & $0.848$ & $0.884$ & $0.897$ & $0.911$ \\
    o1               & $\mathbf{0.885}$ & $\mathbf{0.952}$ & $\mathbf{0.963}$ & $\mathbf{0.970}$ \\
    o3-mini          & $0.865$ & $0.905$ & $0.918$ & $0.931$ \\
    DeepSeek-V3      & $0.826$ & $0.888$ & $0.901$ & $0.911$ \\
    DeepSeek-R1      & $0.741$ & $0.796$ & $0.813$ & $0.837$ \\
    Qwen-Max         & $0.767$ & $0.831$ & $0.847$ & $0.861$ \\
    Claude-3.7       & $0.869$ & $0.922$ & $0.930$ & $0.941$ \\
    Claude-4       & $0.880$ & $0.912$ & $0.921$ & $0.931$ \\
    \midrule
    Human Performance & $0.961$ & N/A    & N/A    & N/A    \\
    \midrule[1pt]
    \multicolumn{5}{l}{\textbf{Canonical Dataset}} \\
    \midrule
    Model Name       & Valid @1 & Valid @3 & Valid @5 & Valid @10 \\
    \midrule
    GPT-4o-mini      & $0.635$ & $0.719$ & $0.750$ & $0.782$ \\
    GPT-4o           & $0.789$ & $0.905$ & $0.931$ & $0.957$ \\
    GPT-4            & $0.849$ & $0.947$ & $0.963$ & $0.973$ \\
    GPT-3.5-Turbo    & $0.714$ & $0.844$ & $0.870$ & $0.886$ \\
    o1-mini          & $0.699$ & $0.847$ & $0.885$ & $0.920$ \\
    o1               & $0.892$ & $\mathbf{0.970}$ & $\mathbf{0.979}$ & $\mathbf{0.987}$ \\
    o3-mini          & $0.855$ & $0.916$ & $0.930$ & $0.936$ \\
    DeepSeek-V3      & $0.787$ & $0.889$ & $0.914$ & $0.930$ \\
    DeepSeek-R1      & $0.867$ & $0.931$ & $0.948$ & $0.961$ \\
    Qwen-Max         & $0.615$ & $0.729$ & $0.775$ & $0.808$ \\
    Claude-3.7       & $0.771$ & $0.899$ & $0.931$ & $0.949$ \\
    Claude-4       & $\mathbf{0.911}$ & $0.950$ & $0.956$ & $0.963$ \\
    \midrule
    Human Performance & $0.951$ & N/A    & N/A    & N/A    \\
    \bottomrule
  \end{tabular}
\end{table}

\begin{table}[h]
  \centering
  \caption{Success@k results of generating targeted generators in two datasets, respectively.}
  \label{tab:hack}
  \begin{tabular}{@{}l|cccc@{}}
    \toprule
    \multicolumn{5}{l}{\textbf{NOIP Dataset}} \\
    \midrule
    Model Name      & Success @1 & Success @3 & Success @5 & Success @10 \\
    \midrule
    GPT-4o-mini     & $0.390$    & $0.610$    & $0.695$    & $0.786$      \\
    GPT-4o          & $0.526$    & $0.755$    & $0.833$    & $\mathbf{0.916}$ \\
    GPT-4           & $0.444$    & $0.679$    & $0.768$    & $0.864$      \\
    GPT-3.5-Turbo   & $0.293$    & $0.558$    & $0.676$    & $0.800$      \\
    o1              & $0.589$    & $0.746$    & $\mathbf{0.797}$    & $0.853$      \\
    o1-mini         & $0.564$    & $0.734$    & $0.785$    & $0.831$      \\
    o3-mini         & $0.627$    & $0.755$    & $0.793$    & $0.839$      \\
    DeepSeek-V3     & $0.422$    & $0.563$    & $0.618$    & $0.670$      \\
    DeepSeek-R1     & $\mathbf{0.667}$    & $\mathbf{0.756}$    & $0.781$    & $0.800$      \\
    Qwen-Max        & $0.360$    & $0.604$    & $0.706$    & $0.811$      \\
    Claude-3.7       & $0.399$ & $0.568$ & $0.637$ & $0.720$ \\
    Claude-4       & $0.378$ & $0.532$ & $0.594$ & $0.654$ \\
    \midrule
    Human Performance & $0.901$    & N/A        & N/A        & N/A       \\
    \midrule[1pt]
    \multicolumn{5}{l}{\textbf{Canonical Dataset}} \\
    \midrule
    Model Name           & Success @1 & Success @3 & Success @5 & Success @10 \\
    \midrule
    GPT-4o-mini          & $0.401$    & $0.570$    & $0.643$    & $0.715$      \\
    GPT-4o               & $0.421$    & $0.617$    & $0.689$    & $0.766$      \\
    GPT-4                & $0.415$    & $0.615$    & $0.692$    & $0.773$      \\
    GPT-3.5-Turbo        & $0.382$    & $0.586$    & $0.668$    & $0.756$      \\
    o1                   & $0.480$    & $0.625$    & $0.675$    & $0.719$      \\
    o1-mini              & $0.455$    & $0.629$    & $0.686$    & $0.743$      \\
    o3-mini              & $\mathbf{0.492}$    & $0.604$    & $0.641$    & $0.682$      \\
    DeepSeek-V3          & $0.423$    & $0.605$    & $0.668$    & $0.719$      \\
    DeepSeek-R1          & $0.441$    & $0.581$    & $0.639$    & $0.710$      \\
    Qwen-Max             & $0.364$    & $0.572$    & $0.659$    & $0.754$      \\
    Qwen2.5-14B          & $0.283$    & $0.431$    & $0.491$    & $0.567$      \\
    Qwen2.5-14B Finetuned& $0.402$    & $\mathbf{0.696}$    & $\mathbf{0.811}$    & $\mathbf{0.917}$      \\
    Claude-3.7       & $0.380$ & $0.524$ & $0.585$ & $0.667$ \\
    Claude-4       & $0.321$ & $0.447$ & $0.500$ & $0.557$ \\
    \midrule
    Human Performance    & $0.821$    & N/A        & N/A        & N/A       \\
    \bottomrule
  \end{tabular}
\end{table}

\begin{table}[h]
    \centering
    \caption{Success@k results of generating targeted generators in the target instruction dataset.}
    \label{tab:noip_with_instrns}
    \begin{tabular}{@{}l|cccc@{}}
        \toprule
        Model Name     & Success @1 & Success @3 & Success @5 & Success @10 \\
        \midrule
        GPT-4o-mini    & $0.680$    & $0.796$    & $0.827$    & $0.859$     \\
        GPT-4o         & $0.748$    & $0.855$    & $0.891$    & $\mathbf{0.931}$     \\
        GPT-4          & $0.677$    & $0.775$    & $0.806$    & $0.841$     \\
        GPT-3.5-Turbo  & $0.583$    & $0.732$    & $0.776$    & $0.817$     \\
        o1             & $0.790$    & $0.821$    & $0.832$    & $0.850$     \\
        o1-mini        & $0.746$    & $0.868$    & $\mathbf{0.896}$    & $0.915$     \\
        o3-mini        & $\mathbf{0.832}$    & $\mathbf{0.878}$    & $0.892$    & $0.898$     \\
        DeepSeek-V3    & $0.742$    & $0.827$    & $0.849$    & $0.864$     \\
        DeepSeek-R1    & $0.779$    & $0.856$    & $0.868$    & $0.879$     \\
        Qwen-Max       & $0.771$    & $0.849$    & $0.867$    & $0.897$     \\
        Claude-3.7       & $0.685$ & $0.79$ & $0.821$ & $0.849$ \\
        Claude-4       & $0.665$ & $0.765$ & $0.799$ & $0.833$ \\
        \bottomrule
    \end{tabular}
\end{table}

\newpage

\end{document}